\documentclass[doc,floatsintext,10pt]{apa6} 

\usepackage[utf8]{inputenc}
\usepackage[american]{babel}
\usepackage{csquotes}
\usepackage[backend=bibtex,style=nature,doi=true,isbn=false,url=false]{biblatex}
\DeclareLanguageMapping{american}{american-apa} 
\usepackage{soul,color}
\usepackage{tipa}
\usepackage{mathtools}
\usepackage{amsmath}
\usepackage{amssymb}
\usepackage{graphicx}
\usepackage{hyperref}
\usepackage{booktabs}

\title{Infant-directed speech is consistent with teaching}
\shorttitle{Infant-directed speech is consistent with teaching}

\fourauthors{Baxter S. Eaves Jr}{Naomi H. Feldman}{Thomas L. Griffiths}{Patrick Shafto}
\fouraffiliations{Rutgers University--Newark}{University of Maryland}{University of California, Berkeley}{Rutgers University--Newark}


\authornote{This work was funded in part by NSF CAREER award DRL-1149116.}

\abstract{
  Infant-directed speech (IDS) has distinctive properties that differ from adult-directed speech (ADS). Why it has these properties -- and whether they are intended to facilitate language learning -- is matter of contention. We argue that much of this disagreement stems from lack of a formal, guiding theory of how phonetic categories should best be taught to infant-like learners. In the absence of such a theory, researchers have relied on intuitions about learning to guide the argument. We use a formal theory of teaching, validated through experiments in other domains, as the basis for a detailed analysis of whether IDS is well-designed for teaching phonetic categories. Using the theory, we generate ideal data for teaching phonetic categories in English. We qualitatively compare the simulated teaching data with human IDS, finding that the teaching data exhibit many features of IDS, including some that have been taken as evidence IDS is not for teaching. The simulated data reveal potential pitfalls for experimentalists exploring the role of IDS in language learning. Focusing on different formants and phoneme sets leads to different conclusions, and the benefit of the teaching data to learners is not apparent until a sufficient number of examples have been provided. Finally, we investigate transfer of IDS to learning ADS. The teaching data improves classification of ADS data, but only for the learner they were generated to teach; not universally across all classes of learner. This research offers a theoretically-grounded framework that empowers experimentalists to systematically evaluate whether IDS is for teaching.}

\begin{document}

\maketitle

Children learn language from input, but often the input children receive differs markedly from normal speech. Infant-directed speech (IDS,  also known as ``motherese'') is characterized by reduced speed, elevated pitch and affect, and unusual prosody. Infants are able to distinguish IDS from normal, adult-directed speech (ADS) and prefer IDS over ADS \parencite{Pegg1992}. Subsequently, researchers have sought to answer why it is that adults speak to children in this unusual way. Seminal work by \textcite{Kuhl1997} found that IDS has unusual formant-level properties. Formants are the representative frequencies of vowel phonemes and manifest as peaks in the spectral envelope. The first formant is the lowest frequency peak, the second formant is the second lowest, and so on. 
When plotted in formant space, the vowels that form the ``corners'' of the space of possible vowels (/\textipa{A}/, as in pot; /\textipa{i}/, as in beet; /\textipa{u}/, as in boot) are hyper-articulated, making them more different from one another. This results in an expansion of the vowel space. Intuitively speaking, hyper-articulation should improve the learnability of vowel categories. All things being equal, example clusters that are more distant are easier to identify. This sparked the idea that IDS is for teaching; an idea that after nearly two decades remains a matter of controversy among researchers. 

Research suggests that corner vowel hyper-articulation is not simply an unintended consequence of highly-affectual speech. Corner vowel hyper-articulation is present in speech to infants but not speech to pets \parencite{Burnham2002}. Additionally, corner vowel hyper-articulation is found in speech to foreigners \parencite{Uther2007}, which, outwardly, sounds more like normal, adult speech. In fact, the social learning literature refers to IDS as an \emph{ostensive cue}: a social cue that engages stricter learning mechanisms in its target \parencite[][]{Gergely2007}. 
It would appear that IDS and its unique features are optimized to teach learners the vowel categories of their language.

However, recent work has discovered statistical features of IDS that are potentially detrimental to learning. Other, non-corner vowels are hypo-articulated (closer together) in IDS \parencite{Kirchhoff2005, Alejandrina2013} and within-phoneme variability increases for some vowels \parencite{DeBoer2003, McMurray2013}. Hypo-articulation is argued to be detrimental to learning because clusters of examples become less distinct as they become nearer. Increased variability is argued to be detrimental because as clusters increase in size, their effective borders shrink or overlap, which makes them less discriminable. Additionally, \textcite{Martin2015} found that temporally sequential pairs of vowel phonemes are less discriminable in IDS than in ADS. It would appear that IDS and its unique features may make learning phonetic categories more difficult.\footnote{Related but orthogonal work suggests that infant- and child-directed speech is less intelligible to adults \parencite{bard1983unintelligibility,bard1994unintelligibility}.}

Over the course of the debate about the role of IDS in language learning, researchers have attempted to quantitatively evaluate the benefit of IDS to learners by comparing the outcome of different learning algorithms given IDS and ADS data \parencite{DeBoer2003, Kirchhoff2005, McMurray2013}. These studies have achieved mixed results. \textcite{DeBoer2003} found that a mixture model trained using the expectation-maximization algorithm was better able to recover the means of IDS corner vowel categories from IDS data than it was to recover the means of ADS corner vowel categories from ADS data. \textcite{Kirchhoff2005} explored the usefulness of IDS to training Bayesian automatic speech recognition systems (ASR), finding that the IDS-trained ASR classified certain types of data more effectively than ADS-trained ASR and other types more poorly. \textcite{McMurray2013} found that multinomial logistic regression trained on IDS data correctly classified fewer new IDS examples than its ADS-trained counterpart classified new ADS examples. Based on these results, the debate appears only to be farther from being resolved. 

We argue that much of the disagreement in the literature with respect to whether IDS is optimized for teaching stems from a lack of a coherent theoretical framework for characterizing teaching. In the absence of such a framework, researchers have substituted intuitions about learning. This has three significant limitations. First, researchers have largely intuited which qualitative features are desirable and which are not. Second, existing computational approaches have attempted to assess teaching indirectly through improvements in learning using various, very different, computational models. Moreover, assessments of model performance have not focused on the key question: the implications of training on IDS for categorization of ADS. Third, the literature tends to focus attention on subsets of the data, both in terms of the vowels and the formants considered for any given analysis. 

Each limitation potentially undermines interpretation.
First, computational models are preferable to intuitive arguments precisely because intuition is fallible, especially when considering the kinds of interactions involved in teaching many categories in a low-dimensional space. Second, while we would expect teaching to lead to better learning, teaching is defined in terms of the intent of the speaker, thus improvements in learning are not a necessary implication---especially if the learner used for performance benchmarking solves a different problem than the learner for whom the teacher generates data. Moreover, given that learners ultimately need to acquire ADS, any improvements in learning should be in transfer between IDS and ADS. Third, because teaching involves considering not just the target vowel but also potentially confusable alternatives, any results derived from subsets of the data may lead to unrepresentative predictions. It is thus important to investigate whether these limitations do affect conclusions in the literature.

Our contribution to the debate is a formal theoretical analysis of how phonetic categories should optimally be taught to infant-like learners. This is the first work to directly address whether IDS is consistent with optimal teaching. We begin by defining the teaching and learning problems under a probabilistic framework. From this model, we generate data designed to teach. We address whether certain features of data are consistent with teaching by qualitatively comparing the features of the teaching data with those of IDS. We address whether IDS-like data are beneficial for learning normal (ADS) speech, and whether these effects generalize, by comparing learning transfer under the target learning model and under standard machine learning algorithms. We also identify some important caveats related to computational analyses based on subsets of data. We address the problems with looking at dimensional and categorical subsets of the data by comparing the features of, and learning outcomes given the original teaching data with those of the teaching data projected onto two-formant space, and we compare the effect of sample size (the number of IDS examples) on learning performance given ADS data and teaching data. We conclude by discussing limitations of the current work and future directions.

\section{Teaching and learning}

To simulate teaching, we must define the components of teaching. In this section we define, in mathematical terms, the components of the problem: the teacher, the learner, and the concept to be learned and taught. Mathematically defining the concept  (the phonetic category model) is matter of applying a formalism that is sufficiently representative of the concept. Similarly, defining a learner requires applying a learning framework that is capable of learning the concept and does so in a psychologically-valid way. And, as we shall see, defining a teacher requires defining a data selection method that is intended to induce the defined concept in the defined learner. Throughout the paper, the words \emph{teacher} and \emph{learner} will be used to refer to the definitions in this section; we will make the necessary distinction when referring to human learners.

\subsection{What is being taught and what is being learned}

In their work on automatic speech recognition, \textcite{Kirchhoff2005} posed the question of what is being learned from IDS. 
If IDS is for teaching then what does IDS teach? While it is typically implied that the intent would be to teach normal speech, existing computational studies compare the effectiveness of IDS at teaching IDS with the effectiveness of ADS at teaching ADS \parencite{DeBoer2003, McMurray2013}. That is, these studies evaluate whether IDS is better at teaching an abnormal (non-adult) speech model than ADS is at teaching the normal speech model. Here, we assume that it is the intent of a teacher to teach the set of phonetic categories used in normal speech.

Building on previous research formalizing phonetic categories, we adopt a Gaussian mixture model (GMM) framework \parencite{DeBoer2003,Vallabha2007,feldmaninpress, McMurray2009b}. Each phonetic category is represented as a multidimensional Gaussian in formant space. We focus on the first, second, and third formants, denoted F$_1$, F$_2$, and F$_3$,  which we capture with 3-dimensional Gaussians. 

A GMM is defined by the probability density function

\begin{equation}
    f(X|\pi_1,\dots\pi_k, \mu_1,\dots, \mu_k, \Sigma_1,\dots\Sigma_k)
    =
        \sum_{i=1}^{k}
            \pi_i
            \mathcal{N}(X|\mu_i, \Sigma_i),
\end{equation}

\noindent
where $\{\pi_1,\dots, \pi_k\}$ is a set of $k$ components weights (real numbers between 0 and 1 inclusive and which sum to 1), $\{\mu_1,\dots, \mu_k\}$ is a set of component means, $\{\Sigma_1,\dots, \Sigma_k\}$ is the set of component covariance matrices, and $\mathcal{N}(X|\mu, \Sigma)$ is the Normal (Gaussian) probability density function applied to the data $X$ given $\mu$ and $\Sigma$.

Importantly, we view the {\em whole system} of phonetic categories as being the object that is being taught. The best data for teaching a single phonetic category might be different from the best data for teaching that category in the context of a set of other categories. When learning a single category, data that are representative of that category are sufficient to communicate the relevant statistical information. When learning multiple categories, without a clear indication of what category each sound belongs to, the possible ambiguity of each sound interacts with the need to provide good information about the statistics of each category to create a much more complex problem.

\subsection{Learning}
\label{sec:learning}

Teaching data are by definition generated with the learner in mind \parencite{Shafto2008,Shafto2014}. A teacher chooses data to induce the correct belief in learners, hence we must define the learner.

Previous computational accounts of learning under IDS have evaluated learning in computational learners that know the correct number of categories \parencite{DeBoer2003} or learn from labeled data \parencite{McMurray2013}. These approaches miss an important difficulty of the learning problem infants face. Infants are not born knowing how many phonemes comprise their native language nor are they given veridical feedback as to which phonetic categories individual components of utterances belong to.  In order to learn the locations (means, $\mu$) and shapes (covariance matrices, $\Sigma$) of phonetic categories, infants must learn how many there are; all while inferring to which phonetic categories each example belongs.

Learning the nature and the number of categories simultaneously can be done using the Dirichlet process Gaussian Mixture Model (DPGMM) \parencite{Anderson1991,Escobar1995,Rasmussen2000,Sanborn2010}. The basic idea is that when a learner cannot assume a fixed number of categories, she must allow for the possibility that there may be as many categories as there are data. This problem can be addressed by using a probabilistic process that determines which data are assigned to which categories \parencite[see][]{Rasmussen2000}. Rather than learning the weights of infinitely many categories, the learner learns an assignment, $Z=\{z_1, \dots, z_n\}$ where $z_i$ is an integer indicating to which component of the mixture the $i^{\text{th}}$ datum belongs. Imagine that we have observed $n$ examples to which we have attributed $k$ categories. Assuming no upper bound on the number of categories, a new example may be assigned to one of the $k$ existing categories or---if it is especially anomalous---may warrant creation of a new, singleton category (a category of which datum $n+1$ is the only member). The mixture weights are then implicit in $Z$. Components with more assigned data have higher weights. We outline this approach in more detail in \autoref{app:model}.

\subsection{Teaching}
We employ an existing model of teaching that has been used successfully to capture human learning in a variety of scenarios \parencite{Shafto2008, Bonawitz2011, Shafto2014, Gweon2014}, under which optimal teaching data derive from the inverse of the learning process. Rather than sampling data randomly from the true distribution, optimal data for teaching are sampled from the distribution that leads learners to the correct inference. Thus teaching involves directing learners' inferences; not just toward the correct hypothesis, but away from alternatives.

Mathematically, the goal of the teacher is to maximize the posterior probability that the learner ends up with the correct hypothesis---in this case, the correct estimate of the category assignments $Z$ and the mixture parameters $\boldsymbol{\mu}$ (all the means $\mu$) and $\boldsymbol{\Sigma}$ (all the covariance matrices $\Sigma$). To express this idea---and allow for the fact that there will be some stochasticity in teaching---we define the probability that the optimal teacher generates data $X$ to be proportional to the posterior probability of the correct hypothesis given that value of $X$. Formally, 
\begin{equation}
    \label{eqn:manteach}
    P_{\text{opt}}(X|Z, \boldsymbol{\mu}, \boldsymbol{\Sigma})
    =
        \frac{
            P(Z, \boldsymbol{\mu}, \boldsymbol{\Sigma}|X)
        }{
            \int_X
                P(Z, \boldsymbol{\mu}, \boldsymbol{\Sigma}|X)
            dX
        }
\end{equation}

\noindent
where the denominator normalizes the distribution, ensuring that it sums to 1 over all $X$. 

Recall that arguments for or against IDS as pedagogical input in existing research rely on the assumption that the pedagogical intent of data can be measured by its benefit to learners. To the contrary, as we shall see, the benefit of data to learners is not a strict indication of the pedagogical intent of data even in our ideal teacher-learner scenario. For example, if the target concept is complex, large amounts of data may be required before any benefit over random data (data generated directly from the target concept) becomes apparent. Alternatively, the adherence of some data to patterns consistent with pedagogically-selected data does provide evidence of pedagogical intent. But without a rigorous definition of pedagogical data selection one can only guess at what these patterns are.

The output of the teaching model is dependent on what is being taught and how it is being taught. Because our goal is to evaluate a claim in the literature, in keeping with the literature---which is framed in terms of learning phonemes from formants---we generate data to teach a subset of language \parentext{a specific phonetic category model derived from \textcite{Hillenbrand1995}} by manipulating first, second, and third formant values. Formants are known to correlate with vowel identity \parencite{Hillenbrand1995,pet52}, though the dimensions that listeners use when storing and categorizing sounds may be more complex than absolute encoding of formant frequencies \parencite{apf15,mcm11,mon10,pet61}. Listeners' reliance on perceptual dimensions may also change over the course of development \parencite{jus92a,nit04}. Thus, our characterization is a significant simplification of the real-world problem.  It makes the teaching problem both easier and more difficult. It is easier because a less complicated model requires less computation to teach, and a teacher need not be concerned with which features are relevant to learners or whether learners must learn which features are relevant. On the other hand, the task is more difficult because we have reduced the information to the learner and reduced the number of manipulable dimensions for the teacher. Thus, the teaching output should be interpreted with care. Differences between our formalization of the problem and nature's will result in differences between the model output and empirical data. We expect the output to be qualitatively similar to human IDS, but do not expect all observed trends to match exactly.

\section{Comparison with Human Infant-Directed Speech}

To evaluate the predictions that this formal model makes about the optimal data for teaching a system of phonetic categories, we focus on twelve American English vowel phonemes and their first, second, and third formants, F$_1$, F$_2$, and F$_3$. \textcite{Hillenbrand1995} provide 48 examples of each phoneme from female speakers. Examples with unmeasurable formant values were discarded, leaving several phonemes with fewer examples \parentext{see \autoref{tab:phonemes}}. The target model -- the one that teachers should be trying to convey to learners -- was derived from the means and covariance matrices calculated from each phoneme's examples (the full list of phonemes and their means and variances can be found in \autoref{tab:phonemes}).

\begin{table}[htb]
\centering
\caption{List of Phonemes in International Phonetic Alphabet Transcription with Means and Variances Calculated from \textcite{Hillenbrand1995}.}
\resizebox{\textwidth}{!}{\begin{tabular}{c l c c c c c c c c c c}
\multicolumn{3}{c}{} & \multicolumn{3}{c}{mean} & \multicolumn{3}{c}{variance} & \multicolumn{3}{c}{covariance} \\
\cmidrule(r){4-6}\cmidrule(r){7-9}\cmidrule(r){10-12}\\
IPA &   e.g. & n & F$_1$ &   F$_2$ &   F$_3$ &   F$_1$ &    F$_2$ &    F$_3$ & F$_1$-F$_2$ &    F$_1$-F$_3$ &    F$_2$-F$_3$\\
\hline
\textipa{\ae} & bat &  47 & 678.06 & 2332.47 & 2972.68 & 4627.84 & 25475.73 & 40006.61 & -4247.73 & -1274.09 & 21255.98 \\
\textipa{A}   & pot & 47 & 916.36 & 1525.83 & 2822.57 & 8449.84 & 15615.80 & 27556.25 &  4354.50 &  1197.37 &   448.93 \\
\textipa{O}   & bought & 47 & 801.02 & 1188.28 & 2819.21 & 5172.15 & 16614.68 & 44701.74 &  6057.43 &   128.67 &    99.29 \\
\textipa{E}   & bet & 48 & 726.67 & 2062.54 & 2952.35 & 5454.06 & 20402.51 & 36093.30 &  -854.33 &  3539.42 & 11775.23 \\
\textipa{e}   & bait & 44 & 536.86 & 2517.09 & 3049.86 & 3807.70 & 24872.41 & 32855.10 & -1656.22 & -1608.30 & 19084.57 \\
\textipa{\textrhookrevepsilon} & Bert & 40 & 526.60 & 1589.35 & 1929.85 & 2193.73 & 12356.90 & 17234.28 &  -402.32 &   989.35 & 10092.08 \\
\textipa{I}   & bit & 48 & 484.31 & 2369.10 & 3057.12 & 1181.03 & 22330.69 & 36138.92 &  -182.84 &  1726.00 & 19153.52 \\
\textipa{i}   & beet & 45 & 435.47 & 2755.96 & 3372.76 & 1662.21 & 20746.41 & 56255.83 &   967.00 &  1010.07 & 18241.44 \\
\textipa{o}   & boat & 48 & 555.46 & 1035.52 & 2828.29 & 6496.21 & 15020.30 & 35040.38 &  6953.69 &   -16.69 &   771.31 \\
\textipa{U}   & put & 48 & 518.65 & 1228.56 & 2829.44 & 1695.72 & 20907.53 & 33424.00 &  2399.33 &   232.84 &  1976.00 \\
\textipa{2}   & but & 48 & 760.19 & 1415.67 & 2900.92 & 3312.88 & 13318.10 & 29810.38 &  2538.87 &  3730.06 &  6977.70 \\
\textipa{u}   & boot & 48 & 459.67 & 1105.52 & 2735.40 & 1496.06 & 42130.34 & 19576.20 &  -417.93 &   -57.95 &  2436.00 \\
\end{tabular}}
\label{tab:phonemes}
\end{table}

\begin{figure}[htb]
    \fitfigure{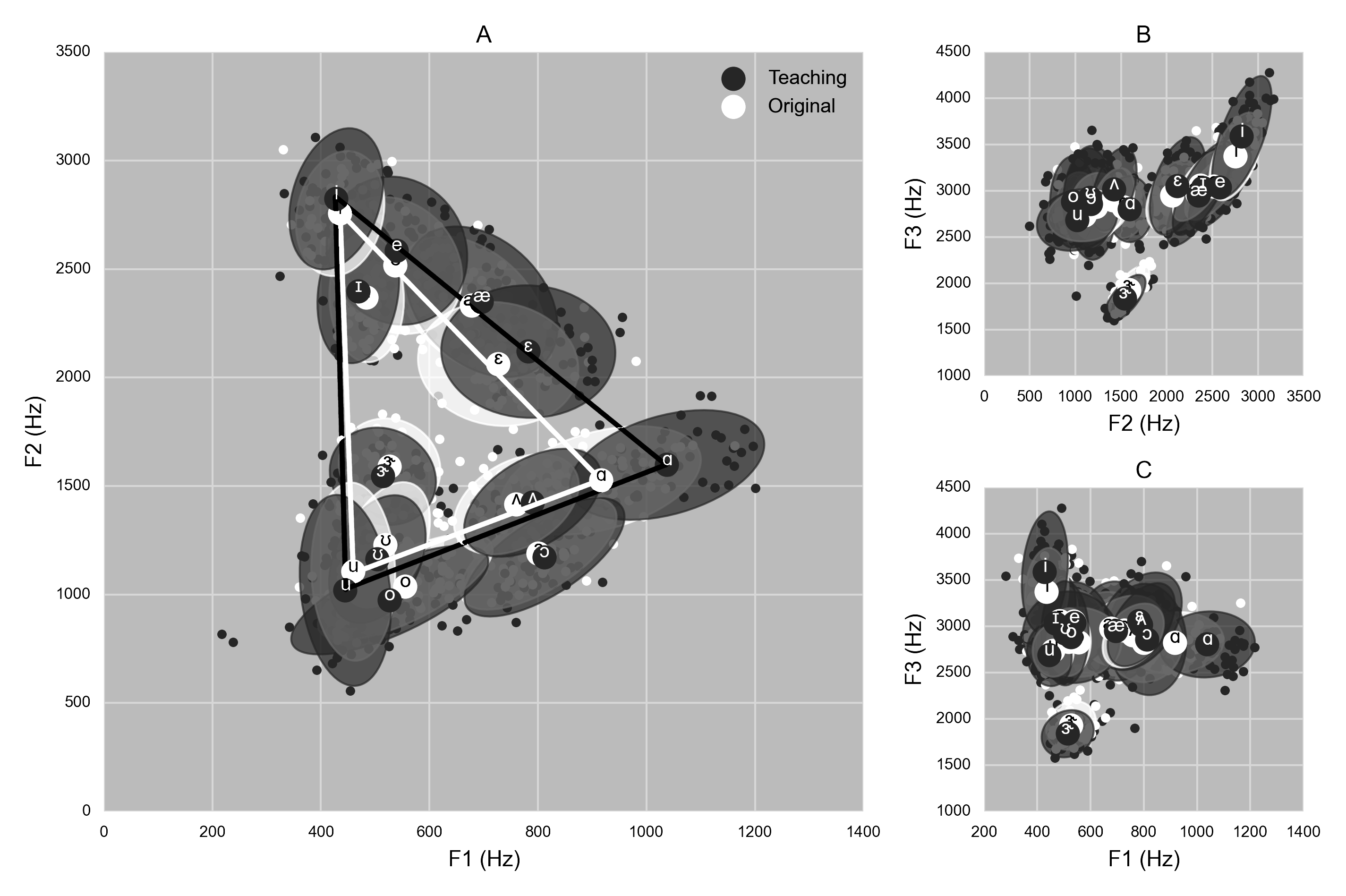} 
    \caption{Distributions of vowels along first, second, and third formants ($F_1$, $F_2$, and $F_3$) in adult-directed speech (light) and speech optimized for the learner (dark). Differences in distributions correspond to the properties of infant-directed speech. Labels are placed at each mean, ellipses represent covariance matrices, and points are a randomly-selected subset of samples from the teaching data and the full set of adult data. All of the original ADS data are represented while a random subset of the teaching data are represented. The light and dark triangles represent the corner vowel triangles for adult-directed and teaching examples, respectively.}
    \label{fig:samples}
\end{figure}

Using an algorithm outlined in \autoref{app:model}, we generated a total of 10,000 samples from the distribution defined in \autoref{eqn:manteach}, each consisting of one example of each of the 12 phonetic categories. We then analyzed these samples, comparing them to human ADS and IDS. \autoref{fig:samples}a shows the distributions of the ADS vowels and the model predictions for IDS along the first and second formants.

\begin{figure}[htb]
    \fitfigure{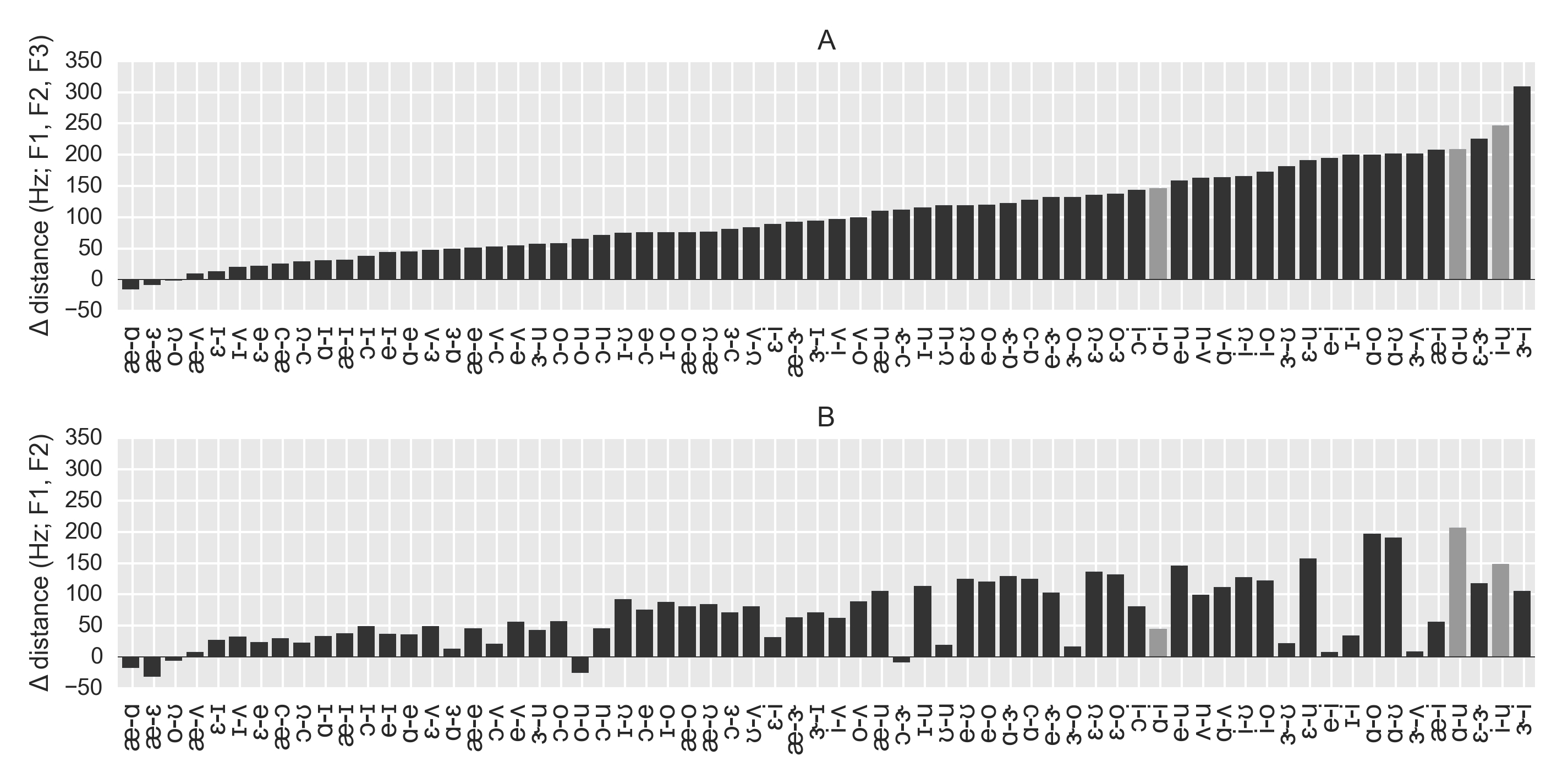}
    \caption{Change in Euclidean distance (Hz; vertical axis) between phonemes pairs (horizontal axis) from ADS to teaching data. Gray bars represent corner vowel pairs. \emph{A}) Given the full, three-formant data. \emph{B}) Given the three-formant data projected onto the F$_1$-F$_2$ plane.}
    \label{fig:articulation}
\end{figure}

\begin{figure}[htb]
    \fitfigure{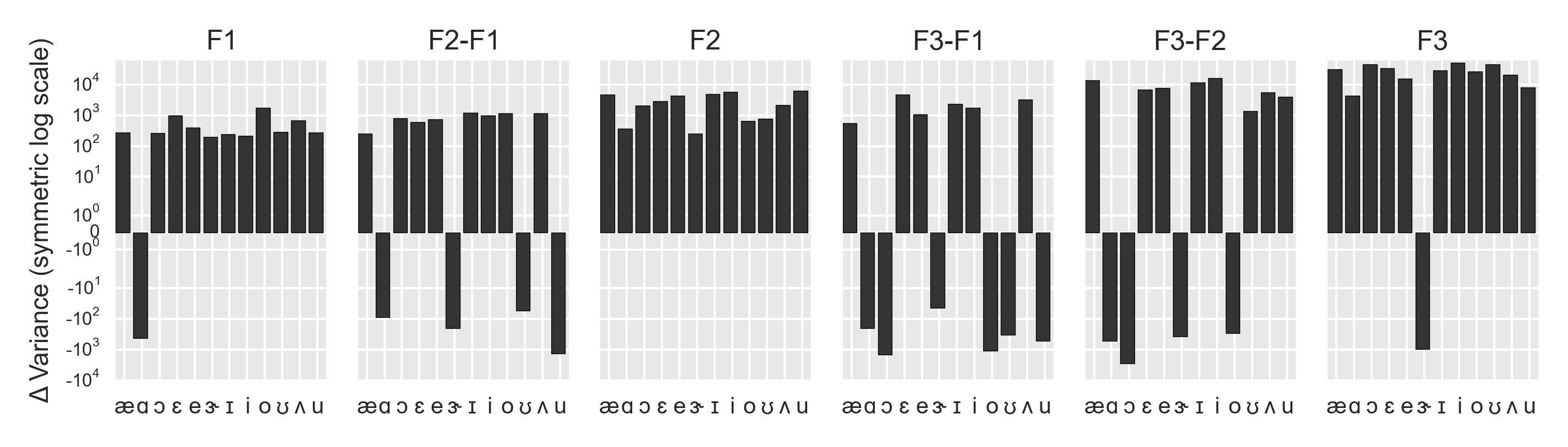}
    \caption{Change in variance, and covariance (symmetric log scale vertical axis) from ADS to teaching data for each phoneme (horizontal axis).}
    \label{fig:variance}
\end{figure}

The model predicts that the simulated teaching data do not simply parrot the target distribution but modify it in ways that match infant-directed speech. Specifically, consistent with previous research \parencite{Kuhl1997,Alejandrina2013,Burnham2002} the corner vowels are hyper-articulated. Additionally, features that researchers have used to argue against the potential pedagogical intent of IDS are present in the teaching data.
\autoref{fig:articulation} shows the predicted change in Euclidean distance between all pairs of vowels. We chose Euclidean distance rather than a variance-based measure of intelligibility because hyperarticulation is defined in terms of movement, and because teaching is meant to communicate the entire category model, not just pairs of phonemes. Most vowel pairs are hyper-articulated, but consistent with IDS, and contrary to previous arguments that IDS is not for teaching \parencite{Alejandrina2013}, the simulated teaching data include hypo-articulation of some vowel pairs. 
\autoref{fig:variance} shows the predicted effects on within-category variability. Consistent with IDS \parencite{DeBoer2003, Alejandrina2013}, but contra previous arguments \parencite{McMurray2013}, the statistically optimal input includes increases in within-category variability for most categories. Of note is the the difference in behaviour between variances and covariances. Other than /\textipa{A}/ in $F_1$ and /\textipa{\textrhookrevepsilon}/ in $F_3$, each phoneme's variance increases. The covariance behavior is less uniform. Four of twelve phonemes decrease $F_1$-$F_2$ covariance, six of twelve decrease $F_3$-$F_1$ covariance, and four of twelve decrease $F_3$-$F_2$ covariance. This suggests that though the teaching data in general exhibit greater variance, orientation plays a role.

It is important to note that trends in hyper- and hypo-articulation change when the three-formant data are flattened onto two dimensions \parentext{\autoref{fig:articulation}a, b}. \autoref{fig:articulation}a shows the change in distance between each phoneme pair in three dimensions (F$_1$, F$_2$, F$_3$) and \autoref{fig:articulation}b shows the change in distance in the same data within the F$_1$-F$_2$ plane.  All corner vowel pairs are hyper-articulated in both sets, but many of the pairs that are hyper-articulated in three-formant space show little change, or are hypo-articulated, in two-formant space. This demonstrates that measures (and thus, conclusions) derived from a dimensional subset of teaching data may provide an incomplete view of the data. For example, it is not appropriate to argue that the data are not for teaching because the /\textipa{o}/-/\textipa{u}/ and /\textipa{O}/-/\textipa{\textrhookrevepsilon}/ pairs are hypo-articulated in the two-formant projection because the data were not generated to teach using only F$_1$ and F$_2$.More broadly, the absolute formant values that are typically analyzed in IDS research differ from the relative formant encodings hypothesized in many perceptual theories.  It has been suggested, for example, that listeners rely on ratios among formants \parencite{mil89,mon10,pet61} or on comparisons of formants among different vowels from the same speaker \parencite{col10,ger68,lob71}.  Inaccurate assumptions about perceptual dimensions can potentially lead to incorrect conclusions about whether IDS is for teaching.

The simulated teaching data include some divergences from human IDS. IDS studies focus on different languages and dialects, and different interior vowels; because the model output is designed to teach an American English phonetic category model, we limit our discussion of systematic deviations to those between the model output and American English IDS. Though the corner vowels hyper-articulate in the teaching data, American English IDS corner vowels hyper-articulate more uniformly \parencite[see][]{Kuhl1997, Alejandrina2013} than the teaching data, which exhibit most hyper-articulation in /\textipa{A}/. In general, the phonemes in the teaching data move away from the interior of the vowel space in the F$_1$-F$_2$ plane, while \textcite{McMurray2013} observed that  /\textipa{\textrhookrevepsilon}/ and /\textipa{\ae}/ moved toward the interior.\footnote{We assume \textcite{McMurray2013} focused on native American English speakers though they only specify that participants were ``from the Ripon, WI area'' and ``all were Caucasian and lived in homes where English was the primary language'' (p. 366).} \textcite{Alejandrina2013} observed that the F$_1$-F$_2$ distance between the /\textipa{i}/-/\textipa{I}/ pair did not change (or hypo-articulated, depending on the measure) from ADS to IDS.  Given these discrepancies, our analysis cannot be taken on its own to provide conclusive evidence that IDS is optimized for teaching.  It does, however, motivate further investigation of previous findings in the literature that have been presented as evidence against IDS serving a teaching function.

\subsection{Effect on learning}

Earlier we argued that the benefit of teaching data is not a strict indication of its pedagogical intent---the implication being that finding that human IDS does or does not improve the performance of some learning algorithm is not, on its own, evidence that IDS is or is not meant to teach. This raises the question of why we should bother investigating learning at all. Certain patterns of learning behavior may be indicative of the presence or absence of pedagogical intent if they are consistent or inconsistent with the predictions of the theory. In this section we venture to identify such patterns. We explore the benefit of the simulated teaching data to several classes of learner, focusing on classification of IDS and ADS data, as well as the effect training on IDS data has on future classification of ADS data. We also investigate how learning performance changes when learning from specific subsets of formants and as a function of sample size.

We first evaluated whether the simulated teaching data, with their unintuitive pedagogical properties, are detrimental to learners' ability to classify example phonemes. We will first evaluate learning performance under several learning models: logistic regression \parencite{McMurray2013}, support vector machines (SVM) with linear kernels, expectation-maximization on Gaussian mixture models (GMM) \parencite{DeBoer2003}, and the Dirichlet process Gaussian Mixture model (DPGMM; the learner model outlined above, and used as the basis for generating the teaching data). We used the scikit-learn \parencite{scikit-learn} implementation for each algorithm except DPGMM, which we implemented using the standard sequential Gibbs sampling algorithm \parencite[][Algorithm 3]{Neal2000} coupled with intermittent split-merge transitions \parencite{Jain2004}, which improves mixing by allowing the Markov Chain to more easily move between modes in the probability distribution.

Each algorithm classified, in batch, random subsets of the teaching data and sets of ADS data randomly generated from the empirical distribution.\footnote{As researchers, we acknowledge that human learning does not happen in batch, but over time from sequential examples. Sequential Monte Carlo \parencite[SMC; see][]{Sanborn2010} algorithms are designed to handle exactly these problems, but to evaluate sequential learning we must make assumptions about the sequence in which examples arrive. In the absence of a reasonable assumption about the order of examples we must marginalize (enumerate and average) over the $N!$ possible orders, which is computationally intractable.} Each set of data consisted of 500 examples of each phoneme (6000 data points total). Each algorithm classified 500 sets of ADS data and 500 sets of teaching data. Logistic regression and SVM, which must first fit a model to labeled data, were provided an identically sized set of different training data and the GMM was provided with the correct number of categories. The DPGMM's prior distribution was identical to the teacher's. The choice of prior is important; the patterns of movement (hyper- and hypo-articulation and variance increase) depend on the prior assumed by the teacher (the teacher chooses data to teach a learner with a certain prior), hence the benefit of patterns of movement to the learner depend on the level of agreement between the teacher' assumed prior and the learner's prior.  We evaluated the DPGMM based on its inferred assignment at the 500$^{th}$ simulation step. We also evaluated the transfer of learning from teaching data to ADS by having each algorithm classify ADS data after having learned a model from teaching data. This \emph{transfer condition} can be thought of as a simulation of the transfer of IDS to ADS. While this has not been evaluated in previous analyses of IDS, it is the critical condition for determining whether IDS helps learners acquire normal speech.

Similarity between each algorithm's inferred category assignments and the correct category assignments was evaluated via the adjusted Rand Index \parencite[ARI, see][]{hubert1985comparing}. The ARI offers a measure of similarity between categorizations in circumstances in which it does not make sense to count the number of correct categorizations (i.e. to count the number of times items with label $z$ are assigned to category $z$). It makes sense to use counting with logistic regression and SVM  because these algorithms fit models given labeled training data and are then used to explicitly label new examples. The GMM, however, is only provided with the number of categories and does not care about their labels; a GMM can label $k$ categories $k!$ different ways. And in addition to not caring about labels, the DPGMM is not guaranteed to have the same number of categories as the true distributions. We use ARI to evaluate all four models.

ARI is provided two partitions of data into categories: the true partition, which is part of the target model; and the inferred partition, which is generated by the learning algorithm. As an example, the partition $[1, 2, 3, 3]$, of four data into three categories implies that datum one belongs to category one, datum two belongs to category two, and data three and four belong to category three. ARI takes on values from -1 to 1 with expected value 0, and assumes the value 1 when the two partitions of stimuli into categories are identical (disregarding labels). For two partitions $\mathbf{U}$ and $\mathbf{V}$ of $N$ data points into $i$ and $j$ categories, ARI is computed as follows:

\begin{equation}
ARI = \frac{\sum_{ij} \binom{n_{ij}}{2} - [\sum_i \binom{a_i}{2} \sum_j \binom{b_j}{2}] / \binom{N}{2}}{\frac{1}{2} [\sum_i \binom{a_i}{2} + \sum_j \binom{b_j}{2}] - [\sum_i \binom{a_i}{2} \sum_j \binom{b_j}{2}] / \binom{N}{2}}.
\end{equation}

 \noindent where $n_{ij}$ is the number of datapoints assigned to $i$ in $\mathbf{U}$ and $j$ in  $\mathbf{V}$, $a_i$ is the sum $ \sum_j n_{ij}$, and $b_j$ is the sum $ \sum_i n_{ij} $. ARI is an adjusted-for-chance version of the Rand Index \parencite{rand1971objective}, which is a normalized sum of the number of pairs of data points that are assigned to the same category in $\mathbf{U}$ and the same category in $\mathbf{V}$, and the number of data points that are assigned to different categories in $\mathbf{U}$ and different categories in $\mathbf{V}$. 

\begin{figure}[htb]
    \fitfigure{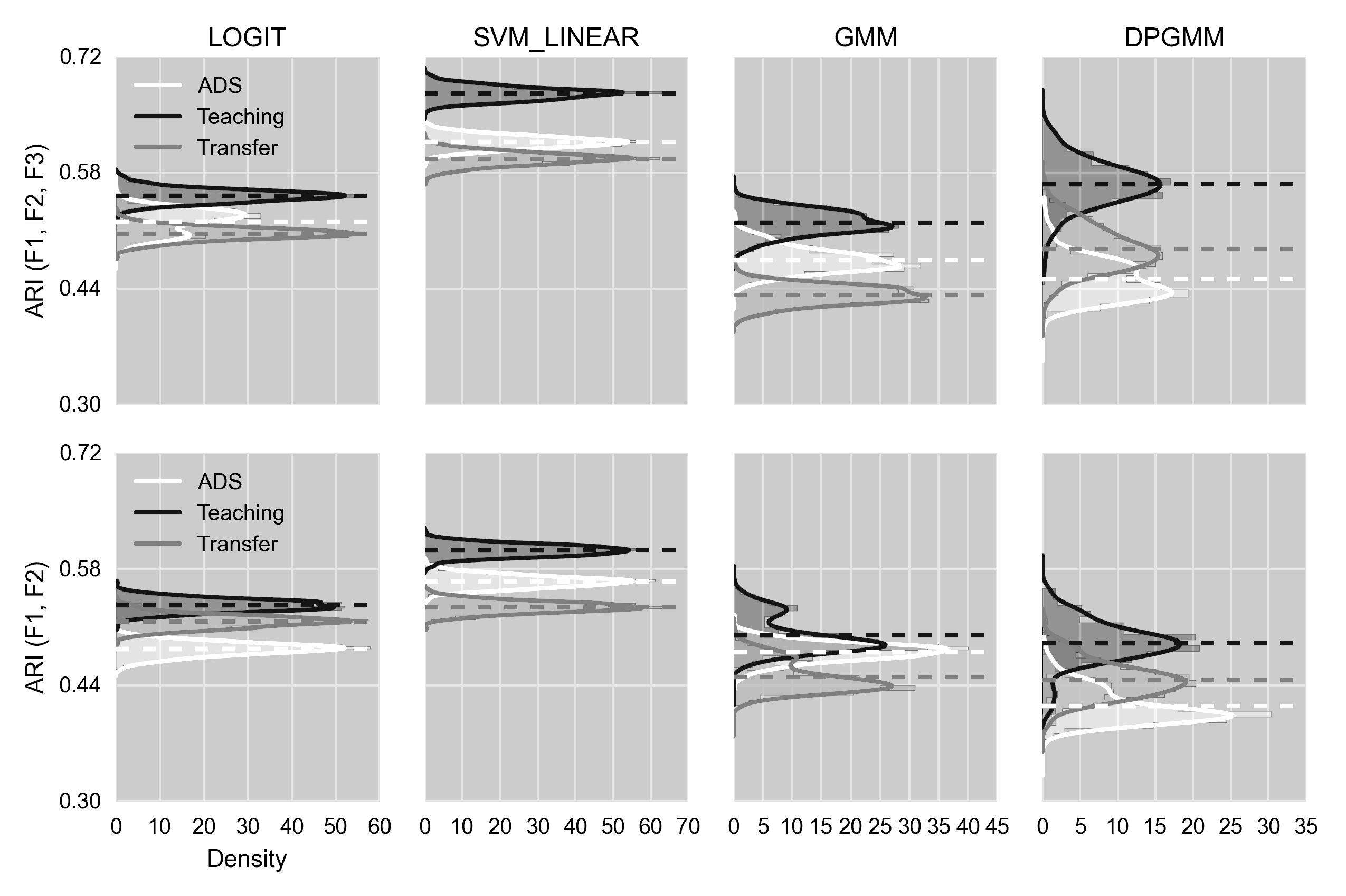}
    \caption{Distributions of ARI for four categorization algorithms (Logistic regression, support vector machine with linear kernel, finite Gaussian mixture model using expectation-maximization, and Dirichlet process Gaussian mixture model) given ADS data generated from the empirical distribution (light), simulated teaching data (dark), and ADS after having learned from teaching data (transfer; gray). \emph{Top row}) ARI given the original, three-dimensional data. \emph{Bottom row}) ARI given the data with the third formant removed.}
    \label{fig:learning}
\end{figure}

 \autoref{fig:learning} (top row) shows that the teaching data (dark) lead to improved classification over ADS (light) data in each of the algorithms we tested. Of the four algorithms, DPGMM performs the worst on the ADS data. This is unsurprising because of the four algorithms, DPGMM has the most to learn. However, DPGMM outperforms GMM on the teaching data. On the full, three-formant data, Logistic regression, SVM, and GMM all perform worst in the transfer condition (gray) compared with the ADS-only and teaching-data-only conditions, while the target learner (DPGMM) classifies ADS data better after having learned from the teaching data. These results show that the teaching data are themselves more classifiable than ADS and improve classification of ADS, in this case, only for the class of learner for which they were intended: the class of learner which must learn the number of phonetic categories. The transfer result is of particular importance and suggests that data that are statistically very different from data generated directly by the true concept can improve learning of the true concept. The real-world implication of this finding is that early learning from IDS may improve future ADS comprehension.

We see that many of the induced ARI distributions in \autoref{fig:learning} are multimodal. Two-sample Kolmogorov-Smirnov (KS) tests indicates that the distribution of ARI given three-formant ADS and teaching data differ under each algorithm; the statistic for each is significant at the $p < 10^{-40}$ level (see \autoref{tab:learning-ks}).\footnote{We use the notation $KS_{LOGIT}(500, 500)=0.668$ to denote that the resulting statistic of a two-sample Kolmogorov-Smirnov test on two samples, both containing 500 data points, equals 0.668} The categorization outcome differs when the three-formant data are projected onto the F$_1$-F$_2$ plane \parentext{see \autoref{fig:learning} bottom row}. Categorization performance generally decreases when F$_3$ is removed. More features (dimensions) provide learners with more information by which they can form categories. For example, in \autoref{fig:samples}b and c we see that locating and categorizing /\textipa{\textrhookrevepsilon}/ (as in Bert) becomes trivial given F$_3$.

\begin{table}[htb]
\centering
\caption{Uncorrected Kolmogorov-Smirnov Test Statistics for \autoref{fig:learning}. Note: $p$ values range from  $\approx10^{-220}$ to $\approx10^{-41}$.}

\begin{tabular}{r l l l l l}
\multicolumn{2}{c}{} & \multicolumn{2}{c}{F$_1$, F$_2$, F$_3$} & \multicolumn{2}{c}{F$_1$, F$_2$} \\
\cmidrule(r){3-4}\cmidrule(r){5-6}\\
Algorithm & Comparison & KS & $p$ & KS & $p$ \\
\hline
Logit & ADS-Teaching      & 0.894 & $\ll 0.0001$ & 0.998 & $\ll 0.0001$\\
      & ADS-Transfer      & 0.584 & $\ll 0.0001$  & 0.972 & $\ll 0.0001$\\
      & Teaching-Transfer & 0.996 & $\ll 0.0001$ & 0.828 & $\ll 0.0001$\\
\hline
SVM (linear) & ADS-Teaching      & 1.0   & $\ll 0.0001$ & 0.994 & $\ll 0.0001$\\
             & ADS-Transfer      & 0.822 & $\ll 0.0001$ & 0.976 & $\ll 0.0001$\\
             & Teaching-Transfer & 1.0   & $\ll 0.0001$ & 1.0   & $\ll 0.0001$\\
\hline
GMM & ADS-Teaching      & 0.872 & $\ll 0.0001$ & 0.434 & $\ll 0.0001$\\
    & ADS-Transfer      & 0.932 & $\ll 0.0001$ & 0.69  & $\ll 0.0001$\\
    & Teaching-Transfer & 1.0   & $\ll 0.0001$ & 0.830 & $\ll 0.0001$\\
\hline
DPGMM & ADS-Teaching      & 0.946 & $\ll 0.0001$ & 0.886 & $\ll 0.0001$\\
      & ADS-Transfer      & 0.54  & $\ll 0.0001$  & 0.596 & $\ll 0.0001$\\
      & Teaching-Transfer & 0.858 & $\ll 0.0001$ & 0.726 & $\ll 0.0001$\\
\end{tabular}
\label{tab:learning-ks}
\end{table}

In the previous paragraphs we demonstrated that the simulated teaching data are indeed beneficial to several classes of learners. It is important to note that these learners benefited from sets of data consisting of a fixed number (500) of examples per phoneme. 

\subsection{Learning as a function of the number of examples}

\begin{figure}[htb]
    \fitfigure{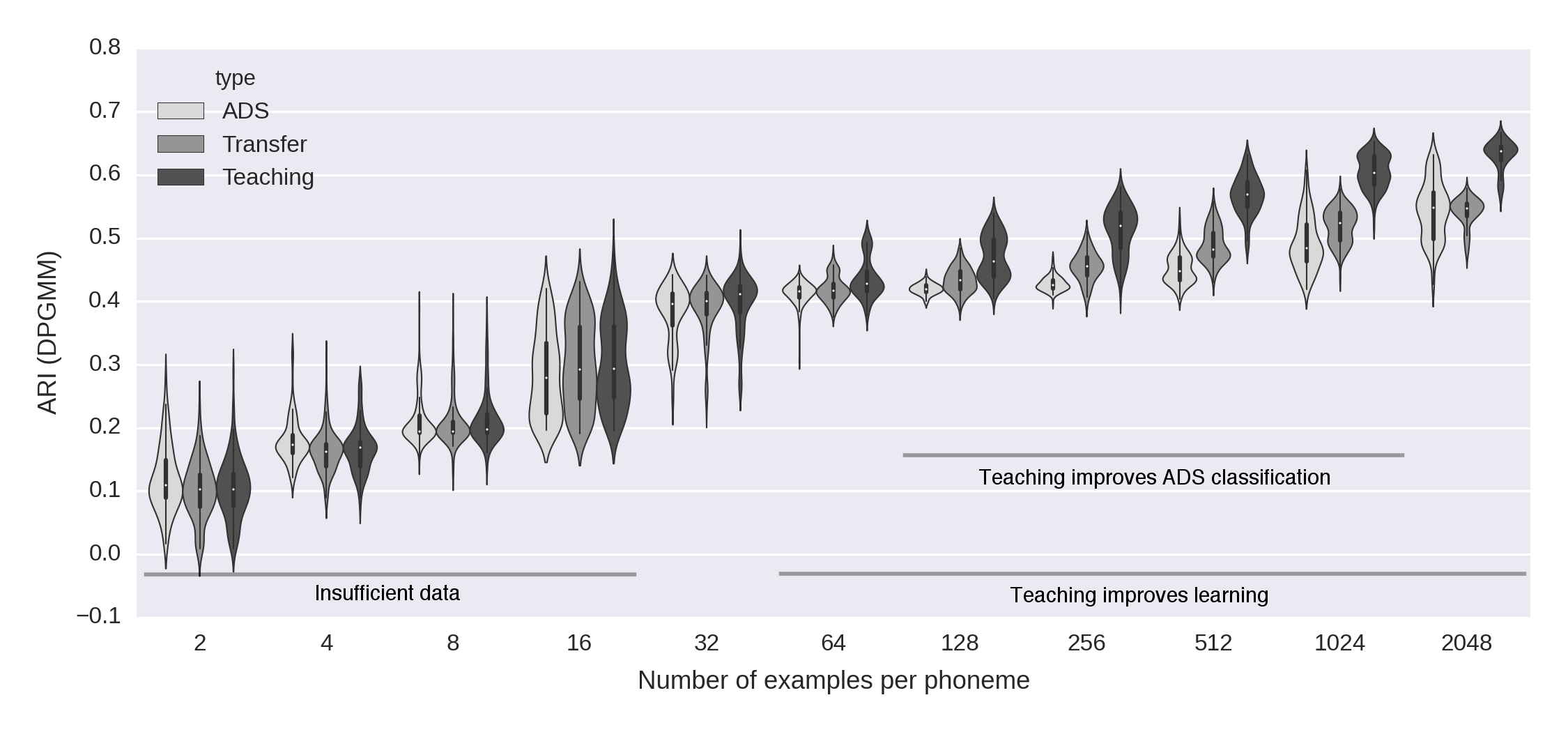}
    \caption{ARI as a function of the number of examples per phoneme for the Dirichlet process mixture model (DPGMM) given ADS data (light), teaching data (dark), and ADS data after learning from teaching data (\emph{transfer}; gray). Low ARI scores at 2, 4, 8, and 16 examples per phoneme indicate that the DPGMM has insufficient data. At 32 examples per phoneme, the teaching data begin to improve learning performance. From 128 to 1024 examples per phoneme, teaching data improve classification of ADS (transfer). }
    \label{fig:ari-over-time}
\end{figure}

Here we investigate how this benefit changes as the number of examples increases or decreases by investigating the effect of the number of examples per phoneme on the classification ability of the target learner (DPGMM). The DPGMM classified 128 random sets of data comprising $2, 4, 8, 16, \dots, 2048$ examples of each phoneme. The results can be seen in \autoref{fig:ari-over-time}. The behavior induced in the DPGMM by the ADS (light) and Teaching (dark) data differ. Adding ADS data appears not to benefit the learner between about 32 and 256 examples per phoneme while adding teaching data continues to improve categorization at an approximately logarithmic rate. This suggest that the benefits of IDS to learners may not be apparent from a small number of data points and that researchers may benefit from comparing learning performance as a function of the number of data points. Learning under ADS begins to improve again after 512 examples, while the benefit of adding ADS examples decreases; and at 2048 examples per phoneme the transfer of IDS results in mean performance similar to ADS. Teaching data are intended to be efficient, thus they should improve learning over random data given a smaller number of examples. If the number of examples is too small, learning is difficult regardless of the data's origin; if the number of examples is sufficiently large, teaching data offer no benefit over random data.

\subsection{Hypoarticulation and increasing variance to teach}
It may be obvious why a teacher would hyper-articulate examples, but the pedagogical usefulness of hypo-articulation and variance increase deserves discussion. 
Keep in mind that the teacher seeks to increase the likelihood of a globally correct inference. 
Hypo-articulation can improve categorization when it is the result of disambiguating movement---that is, movement of one cluster away from another cluster it may be mistaken with. Increased variability can be used to mitigate any negative affects of hypo-articulation by making close or overlapping clusters more distinguishable from each other. Imagine two very closely overlapping, circular clusters: examples from these clusters may appear to come from one large cluster. If we wish to express that there are two clusters we could stretch each cluster perpendicularly so the resulting data manifest as an `X' rather than a single Gaussian blob; indeed, the teaching model produces this behavior. 

The teaching data offer similar examples of how hypo-articulation and increased variability, when employed systematically, do not necessarily reduce learning. For purposes of clarity, we shall look only at the F$_1$-F$_2$ plane (\autoref{fig:samples}a). The phonemes (/\textipa{\textrhookrevepsilon}/; /\textipa{u}/; /\textipa{U}/, as in put; /\textipa{o}/, as in boat) are difficult to distinguish in AD speech. In the teaching data, /\textipa{u}/, /\textipa{U}/, and /\textipa{o}/ are pressed into each other \parentext{hypo-articulated} which makes /\textipa{\textrhookrevepsilon}/ more distinguishable. The corner vowel /\textipa{u}/ greatly increases its F$_2$ variance and decreases its F$_1$-F$_2$ covariance and /\textipa{o}/ greatly increases its F$_1$ variance. This causes /\textipa{o}/ and /\textipa{u}/ to overlap through each other. Their tails then emerge conspicuously from the main mass of examples which makes them more identifiable. The hypo-articulation and directional changes in variance reduce the muddling effect of general increases in within-phoneme variance. Looking at the categorization performance of this subset of the flattened data shows that different algorithms come to different conclusions as to which data are better for learning (we chose categorization results on 500 examples per phoneme). SVM performs better on the ADS data $(M_{ADS}=0.431, M_{Teach}=0.403; KS(500,500)=0.716, p < 0.001; d=2.019)$ and logistic regression performs similarly on ADS and teaching data $(M_{ADS}=0.294, M_{Teach}=0.292; KS(500,500)=0.070, p = 0.166; d=0.109)$. GMM performs better on the teaching data $(M_{ADS}=0.347, M_{Teach}=0.353; KS(500,500)=0.184, p < 0.001; d=-0.301)$, as does DPGMM $(M_{ADS}=0.275, M_{Teach}=0.283; KS(500,500)=0.14, p < 0.001; d=-0.231)$.   These result show first, that hypo-articulation and increased variance do not necessarily damage local inferences in the target model (DPGMM); and second, that looking at categorical subsets of teaching data may lead to conflicting conclusions from different learning algorithms with respect to the benefit of data to learners.

\section{Discussion}

In this paper we have explored the question of whether IDS is for teaching. We rigorously defined both the learning and teaching problems in a psychologically-valid, probabilistic theory. Using this theory, we generated data designed to teach a subset of the phonetic category model of adult speech to naive, infant-like learners using the F$_1$, F$_2$, and F$_3$ formants. In the process, we have identified, concretely demonstrated, and provided possible solutions to a number of issues in the existing literature. We address each in turn. We then conclude by noting the positive results of our analysis, limitations of our results, and recommendations for future research. 

First, the existing literature has relied on intuitive arguments regarding which features of IDS may or may not be desirable. Hyper-articulation (expansion) of the corner vowels has been identified as a feature that would facilitate learning. However, hypo-articulation such as observed between /\textipa{I}/ and /\textipa{i}/ by \textcite{Alejandrina2013}, and increases in variance of categories such as /\textipa{\ae}/  and /\textipa{\textrhookrevepsilon}/ observed by \textcite{McMurray2013}, have been argued to impede learning. Our results show that, when considered in aggregate, hypo-articulation and increases in variance are indeed consistent with teaching. Our analysis leads to predictions about when and why one may see these surprising properties. Hypo-articulation appears when vowels move away from more confusable alternatives. To compensate for this, hypo-articulated categories appear in conjunction with hyper-articulation on other formant dimensions (F$_3$) and/or increases in (co)variance that would facilitate the learner's inference that there is more than one category present. /\textipa{o}/ and /\textipa{u}/ are hypo-articulated in $F_1 \times F_2$, but are hyper-articulated in $F_1 \times F_2 \times F_3$. Both of these phonemes increase their $F_1$ and $F_2$ variance, but /\textipa{o}/ increases its $F_1$-$F_2$ covariance while /\textipa{u}/ decreases its $F_1-F_2$ covariance, which causes the two phonemes to become more conspicuous by overlapping through each other. Thus, our results show that researchers' intuitive theories of which features of IDS are beneficial for teaching are contradicted by a more precise, computational analysis of teaching phoneme categories. 

Second, existing computational approaches have attempted to assess teaching indirectly through improvements in learning using various, very different, computational models and have assessed the benefits of learning from IDS with transfer to IDS. We have argued that the existing models make unreasonable assumptions about the problem faced by the learner. Specifically, models assume that infants either know the number of phonemes in their language \emph{a priori} \parencite{DeBoer2003} or that the data they receive is accompanied by correct labels \parencite{McMurray2013}. Prima facie, these assumptions are too strong. The problem the learner faces includes learning the number of categories. Analyses based on this problem lead to consequential differences in results. Learners who face the problem of learning the number of categories show positive effects of transfer from the simulated teaching data to ADS, while algorithms that assume labeled data or a known number of categories do not (see \autoref{fig:learning}). Our results based on more realistic assumptions about the learning problem contradict previous conclusions in the literature. 

Third, the literature tends to focus attention on subsets of the data, both in terms of the vowels and the formants considered for any given analysis. Both empirical and computational analyses tend to focus on subsets of IDS. Rather than measuring F$_1$, F$_2$ and F$_3$, many analyses rely only on F1 and F2. Similarly, rather than recording data for all vowel categories, results tend to focus on subsets that are relevant to intuitively derived qualitative predictions. Our results show that predictions for teaching depend on knowledge of both of these aspects of context, and thus interpretation of empirical results do as well. As illustrated in \autoref{fig:articulation}, hypo-articulation cannot be determined from F$_1$ and F$_2$ alone; the vowels may be separated on F$_3$. In fact, rhotic vowels such as /\textipa{\textrhookrevepsilon}/ and /\textipa{Ar}/ (as in start) are characterized by low F$_3$ frequencies. Similarly, hypo-articulation may be accompanied by increases in variance, which optimize the learner's ability to infer the existence of more than one category. Thus, our results show that more comprehensive data are necessary to develop accurate computational models and interpret empirical results.

Our results are based on the \textcite{Hillenbrand1995} data, which do not include many of the interior and rhotic vowels use in other studies \parencite{McMurray2013, Alejandrina2013}. Because our results show that quantitative predictions are sensitive to the specifics of context, we do not expect a perfect match to the behavioral data. As we noted, the trends in the simulated teaching data did not exactly match trends others have reported in human IDS. The vowels /\textipa{\textrhookrevepsilon}/ and /\textipa{\ae}/ did not exhibit the interior movement reported by \textcite{McMurray2013}, nor did /\textipa{i}/ and /\textipa{I}/ exhibit F$_1$-F$_2$ hypoarticulation as reported by \textcite{Alejandrina2013}. The qualitative implications of our analysis are more powerful as a consequence: these points illustrate the need for more comprehensive data sets to ensure progress in the debate. 

Building on previous computational models of teaching, we have introduced an approach that may allow direct assessment of whether IDS is intended to teach. The analyses presented here suggest that surprising features identified by researchers are indeed predicted by the model and that IDS is indeed effective for teaching ADS categories provided one assumes a realistic model of learning. Our results also highlight challenges for research investigating the purpose of IDS. 

Implicit in this problem is thus a dependence of teaching data on assumptions of what is being taught. Indeed, this dependence on the set of alternatives is likely what makes desirable features tricky to intuit. If IDS is only for teaching phonetic categories, a more complete set of phonemic data is necessary. 
Though we derived our target phonetic category model from a fairly extensive data set, we hardly encompass the full category model of American English, and may also differ in the perceptual dimensions we assume.\footnote{Additionally, phonemes in \textcite{Hillenbrand1995} were measured only from words beginning with an `h' and ending with a `d' e.g., /\textipa{A}/, /\textipa{i}/, and /\textipa{u}/ were taken only from the words `hod', `heed', and `who'd' respectively.} We lack many of the interior vowels investigated by other researchers \parencite[see][]{Alejandrina2013, McMurray2013}. However, it possible that IDS may be optimized for teaching a larger subset of language. Indeed, research has shown that IDS improves word segmentation \parencite{Thiessen2005}, word recognition \parencite{Singh2009}, and label learning \parencite{GrafEstes2012}. Though daunting, our results highlight the need to systematically consider these alternatives. Our approach, in which we consider categories defined over $F_1$ and $F_2$ versus $F_1$, $F_2$ and $F_3$, can be viewed as a modest start in that direction. With such computational models in hand, it becomes an empirical question, albeit one that requires more comprehensive data than we currently have available. 

Another concern that has not yet been addressed in the literature is differences in learning from individual caregivers and from aggregated data from multiple caregivers. Computational research has sought to answer the question of how people solve inference problems that are computationally intractable, positing that people use approximations \parencite{Sanborn2010}. If this is the case, it is reasonable to assume that different caregivers will arrive at different solutions through stochastic search (e.g. Markov chain Monte Carlo). The distribution of teaching data is highly multi-modal and Markov Chains often find themselves stuck in local maxima. Pilot research suggest data from single chains is far more beneficial to learners than the data aggregated over chains---perhaps due to lower within-phoneme variability compared with aggregated data. We use the aggregated data because it represents the correct probabilistic solution, however because infants are exposed to only a few primary speakers, the literature's tendency to make comparisons over many individuals may misrepresent the problem \parencite[see][for a detailed discussion on how language learners may handle inter-speaker variability]{Kleinschmidt2015}.

This work is also relevant to the articulation literature, where the theoretical underpinnings of speakers' speech manipulations are under debate \parencite[see][]{jagerbuz2016}. The teaching model, coupled with a temporal model of articulation, could predict hyper- or hypo-articulation, and duration increases or decreases. Temporal effects that are explained in terms of a number of heuristics such as planning economy, phonetic neighborhood density, or binary-feature-based addressee-driven attenuation \parencite{Lindblom1990,munson2004effect,galati2010attenuating}, may in fact be consistent with pedagogical manipulation \parentext{as explicitly suggested by \textcite{Lindblom1990, jaeger2013production}}. Related research indicates that speakers adapt when their communications are unsuccessful \parencite{stent2008adapting,schertz2013exaggeration,Buzetalinpress}. However, until the scaling of the teaching model is improved, the problem of temporal articulation will be unapproachable.

\subsection{Conclusion}

Increasingly, research has highlighted ways in which other people may affect learning \parencite{Gergely2002, Koenig2005, Bonawitz2011, Gweon2014}. The problem of language, viewed as statistical learning, is in principle no different. Research has shown that people systematically vary their speech to different targets, with infant directed speech being a canonical example.  It is natural to ask, why. Is it for teaching? We have argued that precise formalization of these hypotheses is a necessary step toward the answer. Building off work in social learning, our computational model of teaching phonemes illustrates limitations in the existing literature. Our approach also points a way forward, through collection of more comprehensive datasets, and development of computational accounts that more accurately reflect the problems faced by learners and hypotheses posited by researchers.

\printbibliography

\appendix
\section{Details of model}\label{app:model}
Here we describe the mathematical details of the model. We construct the teaching model from the learning model.

\subsection{Learner model}
We formalize phonetic category acquisition as learning an infinite Gaussian mixture model \parencite[GMM; see][]{Rasmussen2000,Anderson1991}. A Gaussian mixture model comprises a set of $k$ multidimensional Gaussian components $\theta = \{\{\mu_1, \Sigma_1\}, \dots, \{\mu_k, \Sigma_k\}\}$, where $\mu_j$ and $\Sigma_j$ are the mean and covariance matrix of the $j^{\text{th}}$ mixture component; and an $k$-length vector of mixture weights $\pi=\{\pi_1, \dots, \pi_k\}$, where each $\pi_j$ is a positive real number and the set $\pi$ sums to 1. The likelihood of some data, $X=\{x_i, \dots, x_n\}$, under a GMM is the product of weighted sums,

    \begin{equation}
        P(X|\theta,\pi) = \prod_{i=1}^n \sum_{j=1}^k \pi_j \mathcal{N}(x_i|\mu_i, \Sigma_i),
    \end{equation}

\noindent
where $\mathcal{N}(x|\mu, \Sigma)$ is the Gaussian probability density function applied to $x$ given $\mu$ and $\Sigma$.

We are concerned with the case where the learner infers the assignment of data to categories rather than the component weights. We introduce a length $n$ assignment vector $Z=[z_1, \dots, z_n]$ where $z_i$ is an integer in $1,\dots,k$ representing to which component datum $i$ is assigned. Because the assignment is explicit, we no longer sum over each component. The likelihood is then,

    \begin{equation}
        \label{eqn:gmm-likelihood}
        P(X|\theta,Z) = \prod_{i=1}^n \sum_{j=1}^k \mathcal{N}(x_i|\mu_i, \Sigma_i)\delta_{z_i,j},
    \end{equation}

\noindent
where $\delta_{z_i,j}$ is the Kronecker delta function, which takes the value 1 if $z_i=j$ (data point $x_i$ is assigned to the ${j^{th}}$ category) and the value 0 otherwise.

Learning is then a problem of inferring $\theta$ and $Z$. Prior distributions on individual components, $\{\mu_j, \Sigma_j\}$, correspond to a learner's prior beliefs about the general location ($\mu$), and the size and shape ($\Sigma$) of categories. We assume that $\mu_j$ and $\Sigma_j$ are distributed according to Normal Inverse-Wishart ($\mathcal{NIW}$). Though we made this choice primarily for mathematical convenience, priors of this and similar form have been used successfully to model human behavior \parencite[e.g.][]{feldmaninpress,Kleinschmidt2015}:

\begin{equation}
    \mu_j, \Sigma_j \sim \mathcal{NIW}(\mu_0,\Lambda_0,\kappa_0,\nu_0) \quad \forall\; j\in\{1,\dots,k\},
\end{equation}

\noindent which implies

\begin{equation}
    \Sigma_j \sim \text{Inverse-Wishart}_{\nu_0}(\Lambda_0^{-1}),
\end{equation}
\begin{equation}
    \mu_j | \Sigma_j  \sim \mathcal{N}(\mu_0,\Sigma_k/\kappa_0) \quad \forall\; j\in\{1,\dots,k\},
\end{equation}

\noindent where $\Lambda_0$ is the prior scale matrix, $\mu_0$ is the prior mean, $\nu_0$ is the prior degrees of freedom, and $\kappa_0$ is the number of prior observations. For simulations, we chose vague prior parameters derived from the data.

\begin{equation}\nu_0=3,\end{equation}
\begin{equation}\kappa_0=1, \end{equation}
\begin{equation}\mu_0 = \frac{1}{N}\sum_{i=1}^{N}X_i, \end{equation}
\begin{equation}\Lambda_0 = \frac{1}{K}\sum_{k=1}^{K}\Sigma\left(X_k\right), \end{equation}

\noindent where $\Sigma\left(X_k\right)$ is the empirical covariance matrix of the adult data belonging to category $k$. The prior mean, $\mu_0$, is the mean over the entire data set, and the prior covariance matrix, $\Lambda_0$, is the average of each category's covariance matrix (see \autoref{tab:phonemes}).

To formalize inference over the number of categories, we introduce a prior on the partitioning of data points into components via the Chinese Restaurant Process \parencite{teh2006hierarchical}, denoted $\text{CRP}(\alpha)$, where the parameter $\alpha$ affects the probability of new components. Higher $\alpha$ creates a higher bias toward new components. Data points are assigned to components as follows:
\begin{equation}
    P(z_{i}=j | Z^{-i},\alpha) = 
    \begin{cases} 
        \frac{n_j}{n-1+\alpha} &\mbox{if } j \in 1\dots k \\
        \frac{\alpha}{n-1+\alpha} &\mbox{if } j = k+1 
    \end{cases},
\end{equation}
 
 \noindent where $Z^{-i}$ is $Z$ less entry $i$, $k$ is the current number of components and $n_j$ is the number of data points assigned to component $j$. One is a minimally informative value of $\alpha$ corresponding to a uniform weight over components. 

The standard learning problem involves recovering the true model, defined by $\theta$ and $Z$, from the data, $X$, (give any prior beliefs) according to Bayes' theorem,

\begin{equation}
    \label{eqn:learning}
    P(\theta,Z|X,\mu_0,\Lambda_0,\kappa_0,\nu_0,\alpha)
    =
        \frac{
            P(Z|\alpha)
            P(\theta|\mu_0,\Lambda_0,\kappa_0,\nu_0)
            P(X|\theta,Z)
        }{
            P(X|\mu_0,\Lambda_0,\kappa_0,\nu_0,\alpha)
        }.
\end{equation}

\noindent The Normal Inverse-Wishart prior allows us to calculate the marginal likelihood, $P(X|\mu_0,\Lambda_0,\kappa_0,\nu_0,\alpha)$, analytically \parencite{Murphy2007}, thus, for a small number of data points (the specific number being limited by one's computing power and patience; in our case, the number being thirteen or fewer) we can exactly calculate the above quantity via enumeration. Expanding the terms, the numerator is,

\begin{equation} \label{eqn:numerator}
    P(Z|\alpha)
    \left(
        \prod_{j=1}^k
            \mathcal{NIW}(\mu_j, \Sigma_j|\mu_0,\Lambda_0,\kappa_0,\nu_0)
    \right)
        \prod_{j=1}^k
            \mathcal{N}(\{x_i \in X:Z_i=j\}|\mu_j, \Sigma_j),
\end{equation}

\noindent where the first term, $P(Z|\alpha)$, is the probability of $Z$ under $\text{CRP}(\alpha)$; the second term is the prior probability of the parameters in each component under Normal Inverse-Wishart; and the third term is the (normal) likelihood of the data in each component given the component parameters.

The denominator of \autoref{eqn:learning} is calculable by summing over all possible assignment vectors,  $\{ Z \in \mathfrak{Z}\}$, and integrating over all possible component parameters,

\begin{eqnarray}
    P(X|\mu_0,\Lambda_0,\kappa_0,\nu_0,\alpha)
    &=&
        \sum_{Z\in\mathfrak{Z}} 
            P(Z|\alpha)
            \prod_{j=1}^{k_Z}\iint_{\theta}
                \mathcal{N}(\{x_i \in X:Z_i=j\}|\theta)
                \mathcal{NIW}(\theta|\mu_0,\Lambda_0,\kappa_0,\nu_0)d\theta\\
    &=& 
        \sum_{Z\in\mathfrak{Z}}
            P(Z|\alpha)
            \prod_{j=1}^{k_Z}P(\{x_i \in X:Z_i=j\}|\mu_0,\Lambda_0,\kappa_0,\nu_0),
\end{eqnarray}

\noindent where $k_Z$ is the number of components in the assignment $Z$ and $P(\{x_i \in X:Z_i=j\}|\mu_0,\Lambda_0,\kappa_0,\nu_0)$ is the marginal likelihood of the set of data points in $X$ assigned to component $j$ in $Z$ under a Normal likelihood with Normal Inverse-Wishart prior (this quantity is calculable in closed-form).

\subsection{Teacher model}

Optimal data for teaching are sampled from the distribution that leads learners to the correct inference and away from incorrect inferences \parencite{Shafto2008,Shafto2014}. The teacher must consider the learner's inferences given all possible choices of data. Thus, we normalize over all possible data $X$,

\begin{eqnarray}
    P_{\text{opt}}(X|\theta,Z,\mu_0,\Lambda_0,\kappa_0,\nu_0,\alpha)
    &\propto& 
        \frac{
            P(\theta,Z|X,\mu_0,\Lambda_0,\kappa_0,\nu_0,\alpha)
        }{
            \int_X
                P(\theta,Z|X,\mu_0,\Lambda_0,\kappa_0,\nu_0,\alpha)dX
        },\\
    &=&
        \frac{
            \frac{
                P(Z|\alpha)
                P(X|\theta,Z)
                P(\theta|\mu_0,\Lambda_0,\kappa_0,\nu_0)
            }{
                P(X|\mu_0,\Lambda_0,\kappa_0,\nu_0,\alpha)
            }
        }{
            \int_{X}
                \frac{
                    P(X|\theta,Z)
                    P(\theta|\mu_0,\Lambda_0,\kappa_0,\nu_0)
                    P(Z|\alpha)
                }{
                    P(X|\mu_0,\Lambda_0,\kappa_0,\nu_0,\alpha)
                }
            dX
        }.
\end{eqnarray}

\noindent The term,
\begin{equation}
    P(\theta,Z|X,\mu_0,\Lambda_0,\kappa_0,\nu_0,\alpha)
    = 
        \frac{
            P(X|\theta,Z)
            P(\theta|\mu_0,\Lambda_0,\kappa_0,\nu_0)P(Z|\alpha)
        }{
            P(X|\mu_0,\Lambda_0,\kappa_0,\nu_0,\alpha)
        },
\end{equation}

\noindent is the posterior probability of the true hypothesis given the data---the learner's inference. The learner's inference over alternative hypotheses is captured by the marginal likelihood of the data, $P(X|\mu_0,\Lambda_0,\kappa_0,\nu_0,\alpha)$. The teacher's optimization of the choice of data is captured by the normalizing constant,

\begin{equation}
\label{eqn:teacher}
    \int_X P(\theta,Z|X,\mu_0,\Lambda_0,\kappa_0,\nu_0,\alpha) dX.
\end{equation}

We avoid the need to calculate this quantity directly by sampling from $P_{opt}$ using the Metropolis algorithm \parentext{\textcite{Hastings1970}, see \autoref{app:sampling}} according to the acceptance probability,

\begin{equation}
    A(X'|X) 
    = 
        \text{min}
            \left[
                1, 
                \frac{
                    P(X'|\theta,Z)
                    P(X|\mu_0,\Lambda_0,\kappa_0,\nu_0,\alpha)
                }{
                    P(X|\theta,Z)
                    P(X'|\mu_0,\Lambda_0,\kappa_0,\nu_0,\alpha)
                } 
            \right].
\end{equation}

\section{Algorithm for generating samples}\label{app:sampling}

The normalizing constant in \autoref{eqn:manteach} \parentext{also \autoref{eqn:teacher} in \autoref{app:model}} is analytically intractable. We use the Metropolis-Hastings algorithm to sample from the distribution of teaching data without having to calculate the normalizing constant \parencite{Hastings1970}. The Metropolis-Hastings algorithm can be applied to draw samples from a probability distribution with density $p: x \rightarrow \mathbb{R}^+$ when $p$ can be calculated up to a constant. That is, when there exists a function $f(x)$, where $p(x) = cf(x)$ and $c$ is a constant. A proposal distribution, $q(x'|x)$, is defined that proposes new samples, $x'$, given the current sample, $x$. Beginning with a sample, $x$, a proposed sample, $x'$, is drawn from $q$. The acceptance ratio, $A$, is calculated from $f$ and $q$,
\begin{equation}
    A = \frac{f(x')q(x|x')}{f(x)q(x'|x)}.
\end{equation}

\noindent
It is easy to see that

\begin{equation}
    \frac{f(x')q(x|x')}{f(x)q(x'|x)} = \frac{cf(x')q(x|x')}{cf(x)q(x'|x)} = \frac{p(x')q(x|x')}{p(x)q(x'|x)}.
\end{equation}

\noindent
If $q$ is symmetric, that is $q(x'|x) = q(x|x')$ for all $x, x'$, then $\frac{q(x|x')}{q(x'|x)}$ (the Hastings ratio) cancels from the equation, leaving, 

\begin{equation}
    A = \frac{f(x')}{f(x)},
\end{equation}

\noindent from which we calculate the probability with which $x'$ is accepted,

\begin{equation} 
    P(x'|x) = \text{min}\left[1, A\right].
\end{equation} 

To sample from the distribution of teaching data using the Metropolis algorithm, we calculate the numerator of \autoref{eqn:manteach} exactly via enumeration and propose symmetric Gaussian perturbations to resample data. The acceptance probability is thus,

\begin{equation}
\label{eqn:accrate}
P(X'|X) = \text{min}\left[1, \frac{P(X'|Z, \boldsymbol{\mu}, \boldsymbol{\Sigma})P(X)}{P(X|Z, \boldsymbol{\mu}, \boldsymbol{\Sigma})P(X')} \right].
\end{equation}

\noindent
For the simulations, the sampler simulated one datapoint for each phoneme (twelve total). $X$ comprised twelve data points, one for each phoneme. $X$ was initialized by sampling data from the prior parameters, that is $X_0 \sim N(\mu_0, \Lambda_0/\kappa_0)$ \parentext{see \autoref{app:model}}. At each iteration, new data, $X'$, were generated from $X$ by adding Gaussian noise distributed $N(0,40)$. This proposal distribution was chosen so that the acceptance rate of $X'$ was near the optimal value of 0.23 \parencite[][]{Roberts1997}. $X'$ was then accepted according to \autoref{eqn:accrate}.

The final data comprise samples from 10 independent runs to the sampler. The first 500 samples of each run were discarded, then each 20$^{th}$ sample was collected until 1000 samples had been collected. The full set of data thus contains 10,000 total samples of twelve data points each (one for each of the twelve phonemes) for a total of 120,000 examples. Aggregating data over speakers is common practice in the IDS literature; we conduct analyses on data aggregated over independent runs of the sampler.

\end{document}